\documentclass[11pt,a4paper]{article}
\usepackage[hyperref]{acl2020}
\usepackage{times}
\usepackage{latexsym}

\usepackage{tikz-dependency}
\usepackage{tikz}
\usepackage{tikz-qtree}

\mathchardef\mhyphen="2D 
\newcommand{\transname}[1]{\ensuremath{\mathsf{#1}}}

\newcommand{\sh}{\transname{Shift}}
\newcommand{\re}{\transname{Reduce}}

\newcommand{\nt}{\transname{Non\mhyphen Terminal}}

\newcommand{\fin}{\transname{Finish}}

\usepackage{microtype}

\aclfinalcopy 
\def\aclpaperid{377} 

\title{Enriched In-Order Linearization for Faster Sequence-to-Sequence Constituent Parsing}

\author{Daniel Fern\'{a}ndez-Gonz\'{a}lez \and Carlos G\'{o}mez-Rodr\'{i}guez\\
	Universidade da Coru\~{n}a, CITIC\\
	FASTPARSE Lab, LyS Group \\
Depto. de Ciencias de la Computaci\'{o}n y Tecnolog\'{i}as de la Informaci\'{o}n \\
	Campus de Elvi\~{n}a, s/n, 15071 A Coru\~{n}a, Spain \\
  {\tt d.fgonzalez@udc.es}, {\tt carlos.gomez@udc.es}\\}

\date{}

\begin{document}
\maketitle
\begin{abstract}
Sequence-to-sequence constituent parsing requires a linearization to represent trees as sequences. Top-down tree linearizations, which can be based on brackets or shift-reduce actions, have achieved the best accuracy to date. 
In this paper, we show that these results can be improved by using an in-order linearization instead.
Based on this observation, we implement
an enriched in-order shift-reduce linearization inspired by \citet{Vinyals2015}'s approach, achieving the best accuracy to date on the English PTB dataset among fully-supervised single-model sequence-to-sequence constituent parsers. 
Finally, we apply deterministic attention mechanisms 
to match 
the speed of state-of-the-art transition-based parsers,
thus showing that sequence-to-sequence models can match them, not only in 
accuracy, but also in speed.
\end{abstract}

\section{Introduction}

Sequence-to-sequence (seq2seq) neural architectures have proved 
useful
in several NLP tasks, 
with
remarkable success in some of them such as machine translation, but they lag behind the state of the art in others. 
In constituent parsing,
seq2seq models still 
need to improve
to be competitive in accuracy and efficiency with
their main competitors: transition-based constituent parsers \cite{Dyer2016,Liu2017,nonbinary}.

\citet{Vinyals2015} laid the first stone in seq2seq constituent parsing, proposing a linearization of phrase-structure trees as bracketed sequences following a top-down strategy, which can be predicted from the input sequence of words by any off-the-shelf seq2seq framework. While this approach is very simple, its accuracy and efficiency are significantly behind the state of the art in the fully-supervised single-model scenario.

Most attempts to improve this approach focused on modifying the neural network architecture, while keeping the top-down linearization strategy. As exceptions, \citet{Ma2017} and \citet{LiuS2S17} proposed linearizations based on sequences of transition-based parsing actions instead of brackets. \citet{Ma2017}  
tried
a bottom-up linearization, but they obtained worse results than top-down approaches.\footnote{We also tested empirically that a bottom-up linearization is not suitable for seq2seq parsing and discarded that option.} \citet{LiuS2S17} kept the top-down strategy, but using transitions of the top-down transition system of \citet{Dyer2016} instead of a bracketed linearization, achieving a higher performance.

In transition-based constituent parsing, an in-order algorithm has recently proved superior to the bottom-up and top-down approaches \cite{Liu2017}, but we know of no applications of this approach in seq2seq parsing.

\paragraph{Contributions} In this paper, we advance the understanding of linearizations for seq2seq parsing, and improve the state of the art, 
as follows:
(1) we show that the superiority of a transition-based top-down linearization over a bracketing-based one observed by \citet{LiuS2S17} does not hold when both 
are 
tested
under the same framework.
In fact, we show that the additional information provided by the larger vocabulary in the linearization of \citet{Vinyals2015} is beneficial to seq2seq predictions. (2) We implement a novel in-order transition-based linearization, based on the in-order transition system by \citet{Liu2017}, and manage to notably increase 
parsing accuracy with respect to previous approaches. (3) We enhance the in-order representation of parse trees by adding extra information following the shift-reduce version of the \cite{Vinyals2015} linearization, obtaining state-of-the-art accuracy among seq2seq parsers and on par with some well-known transition-based approaches. (4) We bridge the remaining gap with transition-based parsers - parsing speed - by applying a new variant of deterministic attention
\cite{Kamigaito2017,Ma2017} to restrict the hidden states used to compute the attention vector, doubling the system's speed. The result is a seq2seq parser\footnote{Source code available at \url{https://github.com/danifg/InOrderSeq2seq}.} that, for the first time, matches the speed and accuracy of transition-based parsers implemented under the same neural framework. (5) Using the neural framework of \citet{Dyer2015} as testing ground, we perform a homogeneous comparison among different seq2seq linearizations and widely-known transition-based parsers.

\begin{figure*}
\centering
\small
\begin{tikzpicture}[level distance=0.8cm]
\tikzset{frontier/.style={distance from root=60pt}}
\Tree [.S [.{NP} [.The ] [.public ] ]
[.VP [.is ] [.ADVP [.still ] ] [.ADJP [.cautious ] ] ] [.. ] ]
\end{tikzpicture}
\begin{scriptsize}
\begin{tabbing}
\hspace{1cm}\=\hspace{2.3cm}\= \kill
\> \textbf{Top-down linearizations} 
\>\\[2mm]
\> \ \ a) Bracketed: 
\> \ \ \ \ \ \ \ {\tt ($_S$ ($_{NP}$ XX XX )$_{NP}$ ($_{VP}$ XX ($_{ADVP}$ XX )$_{ADVP}$ ($_{ADJP}$ XX )$_{ADJP}$ )$_{VP}$ XX )$_S$} \\[2mm]
\> \ \ b) Shift-reduce:
\> \ \ \ \ \ \ \ {\tt NT$_S$ NT$_{NP}$ SH SH RE NT$_{VP}$ SH NT$_{ADVP}$ SH RE NT$_{ADJP}$ SH RE RE SH RE} \\[2mm]
\> \ \ c) Enriched SH-RE:
\> \ \ \ \ \ \ \ {\tt NT$_S$ NT$_{NP}$ SH SH RE$_{NP}$ NT$_{VP}$ SH NT$_{ADVP}$ SH RE$_{ADVP}$ NT$_{ADJP}$ SH RE$_{ADJP}$ RE$_{VP}$}\\ 
\> \>\ \ \ \ \ \ \ {\tt SH RE$_S$} \\[2mm]
\> \textbf{In-order linearizations}
\>\\[2mm]
\> \ \ d) Shift-reduce:
\> \ \ \ \ \ \ \ {\tt SH NT$_{NP}$ SH RE NT$_S$ SH NT$_{VP}$ SH NT$_{ADVP}$ RE SH NT$_{ADJP}$ RE RE SH RE FI} \\[2mm]
\> \ \ e) Enriched SH-RE:
\> \ \ \ \ \ \ \ {\tt SH NT$_{NP}$ SH RE$_{NP}$ NT$_S$ SH NT$_{VP}$ SH NT$_{ADVP}$ RE$_{ADVP}$ SH NT$_{ADJP}$ RE$_{ADJP}$ RE$_{VP}$}\\
\> \>\ \ \ \ \ \ \ {\tt SH. RE$_S$ FI}
\end{tabbing}
\end{scriptsize}
\caption{Top-down and in-order linearizations for a constituent tree taken from English PTB. SH = $\sh$, NT$_{\textsc{X}}$ = $\nt$-X, RE = $\re$, RE$_{\textsc{X}}$ = $\re$-X and FI = $\fin$.}
\label{fig:linearizations}
\end{figure*}

\section{Enriched Linearizations}
\label{sec:lin}
To cast constituent parsing as 
seq2seq prediction,
each parse tree needs to be represented as a sequence of symbols that can be predicted from an input sentence.
Initially, \citet{Vinyals2015} proposed a top-down bracketed linearization 
of constituent trees,
where 
opening and closing brackets
include non-terminal labels and POS tags are normalized by replacing them with a tag XX. 
An example is shown
in linearization $a$ of Figure~\ref{fig:linearizations}.

As an alternative, \citet{LiuS2S17} presented a shift-reduce linearization based on the top-down transition system defined for constituent parsing by \citet{Dyer2016} (example $b$ in Figure~\ref{fig:linearizations}). This provides three transitions that can be used on a stack and a buffer to build a constituent tree: a $\sh$ transition to push words from the buffer into the stack, a $\nt$-X transition to push a non-terminal node X into the stack, and a $\re$ transition to pop elements from the stack until a non-terminal node is found and create a new subtree with all these elements as its children, pushing this new constituent into the stack.

Following \citet{Vinyals2015}'s linearization where closing brackets also include the non-terminal label, we define an equivalent shift-reduce variant, where the $\re$ transition is also parameterized with the non-terminal on top of the resulting subtree ($\re$-X). In that way, we can one-to-one map opening brackets to $\nt$-X transitions, closing brackets to $\re$-X actions and XX-tags to $\sh$ transitions as shown in example $c$ of Figure~\ref{fig:linearizations} . This enriched version will enlarge the vocabulary, 
but will also add some extra information that, as we will see below, improves parsing accuracy.

As an alternative to the top-down 
parser of \cite{Dyer2016},
\citet{Liu2017} define a transition system 
based on in-order traversal, as in left-corner parsing \citep{Rosenkrantz70}:
the non-terminal node on top of the 
tree being built
is only considered after the first child 
is completed in the stack, building each subtree in a bottom-up manner, but choosing the non-terminal node on top before the new constituent is reduced. 
Transitions are the same as in
the top-down algorithm (plus a $\fin$ transition to terminate the 
parsing process), but the effect of applying a $\re$ transition is different: it pops all elements from the stack until the first non-terminal node is found, which is also popped together with the 
preceding element
in the stack to build a new constituent with all of them as children of the non-terminal node.\footnote{See Appendix~\ref{sec:appendix} for more details about the top-down and in-order transition systems.}

This algorithm pushed state-of-the-art accuracies in
shift-reduce constituent parsing; and, as we show in Section~\ref{sec:results}, it can be succesfully applied as a linearization method for seq2seq 
constituent 
parsing. Sequence $d$ in Figure~\ref{fig:linearizations} 
exemplifies
in-order linearization.

Similarly to the enriched top-down variant, we 
also
extend the in-order shift-reduce linearization by parametrizing $\re$ transitions. Additionally, we can also add extra information to $\sh$ transitions. \cite{Suzuki2018} leaves POS tags of punctuation symbols out of the normalization proposed by \citet{Vinyals2015} without further explanation, but possibly they consider it can help seq2seq models. We adapt this idea to our novel enriched in-order linearization and lexicalize $\sh$ transitions when a ``.'' or a ``,'' are pushed into the stack as ``$\sh.$'' and ``$\sh,$'', respectively.\footnote{We do not lexicalize $\sh$ transitions on the enriched shift-reduce top-down variant to perform a fair comparison against the original linearization by \citet{LiuS2S17}.} In our experiments, we see that lexicalizing $\sh$ transitions has indeed an impact on parsing performance. In Figure~\ref{fig:linearizations} and sequence $e$, we include an example of this linearization technique.

Note that, although we use a transition-based linearization of parse trees, our approach is agnostic to the stack structure and the 
parsing process is performed by a simple seq2seq model that straightforwardly \textit{translates} input sequences of words into sequences of shift-reduce actions.

\section{Seq2seq Neural Network}
\paragraph{Baseline Model} In our experiments, we test all proposed linearizations in the 
seq2seq neural 
architecture designed by \citet{LiuS2S17} and implemented on the framework developed by \citet{Dyer2015}. This 
architecture proved to outperform the majority of seq2seq approaches, even 
without implementing beam search (which penalizes parsing speed).
The difference with respect to the vanilla seq2seq configuration \cite{Vinyals2015} is that two separate attention models are used to cover two different and variable segments of the input. This provides improvements in accuracy, regardless of the linearization method used.

\begin{figure*}[t]
\centering
\includegraphics[width=0.9\textwidth]{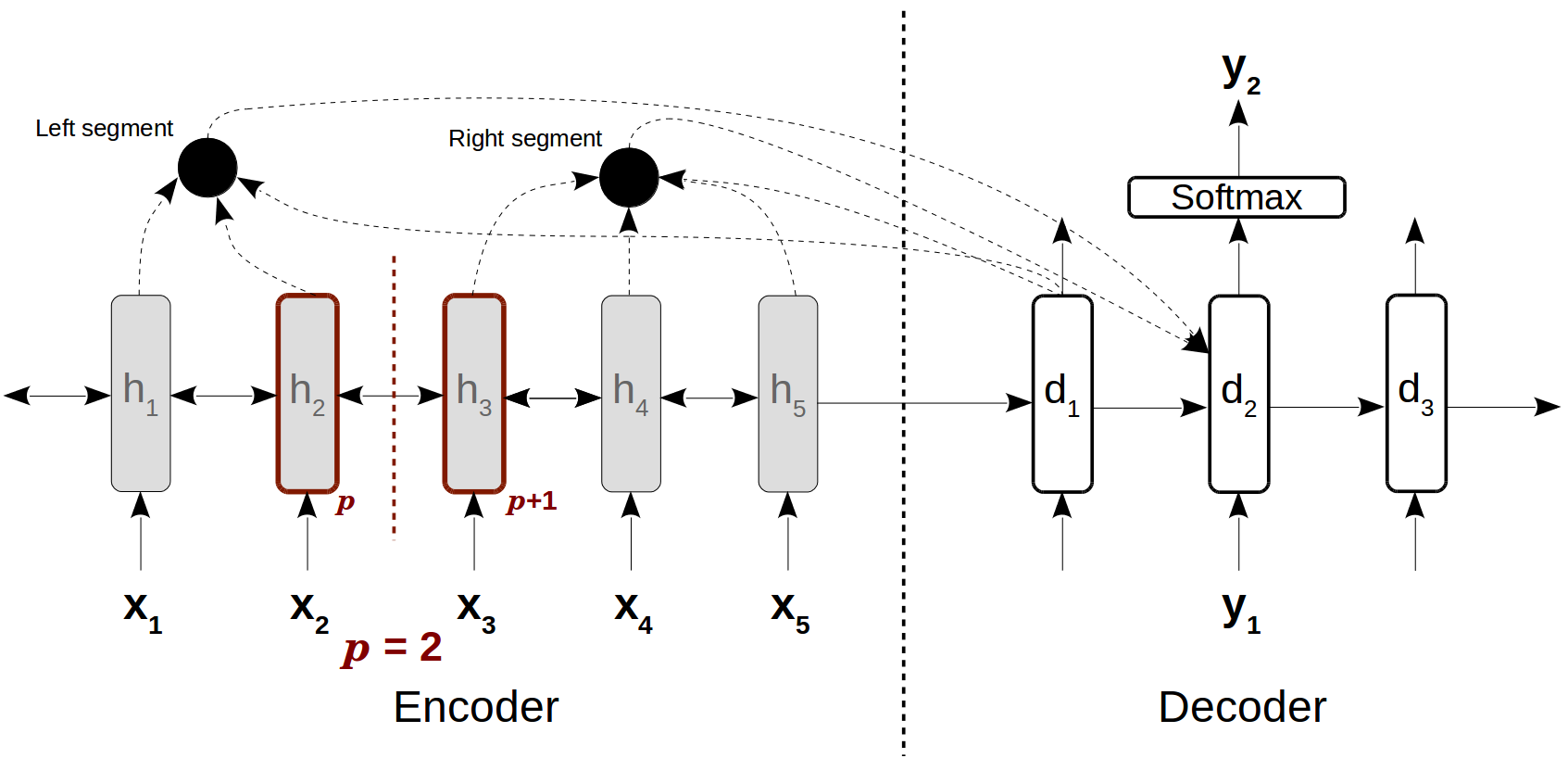}
\caption{Sequence-to-sequence neural architecture proposed by \citet{LiuS2S17}.}
\label{fig:network}
\end{figure*}

More specifically, \citet{LiuS2S17} follow the common practice in stack-LSTM-based shift-reduce parsers \cite{Dyer2015,Dyer2016,Liu2017} that uses a concatenation of pretrained word embeddings ($\mathbf{e}^{*}_{w_i}$) and randomly initialized word ($\mathbf{e}_{w_i}$) and POS tag embeddings ($\mathbf{e}_{p_i}$) to derive (through a ReLu non-linear function) the final representation $\mathbf{x}_i$ of the $i$th input word:
$$\mathbf{x}_i = \mathbf{relu}(\mathbf{W}_{enc}[\mathbf{e}^{*}_{w_i}, \mathbf{e}_{w_i}, \mathbf{e}_{p_i}] + \mathbf{b}_{enc})$$
\noindent where $\mathbf{W}_{enc}$ and $\mathbf{b}_{enc}$ are model parameters, and $w_i$ and $p_i$ represent the form and the POS tag of the $i$th input word.

This representation $\mathbf{x}_i$ is fed into the encoder (implemented by a BiLSTM) to output an \textit{encoder hidden state} $\mathbf{h}_i$:
$$ \mathbf{h}_i = [\mathbf{h}_{l_i};\mathbf{h}_{r_i}] = \mathbf{BiLSTM}(\mathbf{x}_i).$$

As a decoder, a LSTM generates a sequence of \textit{decoder hidden states} from which a sequence of actions is predicted. Concretely, the current decoder hidden state $\mathbf{d}_j$ is computed by:
$$\mathbf{d}_j = \mathbf{relu}(\mathbf{W}_{dec}\big[\mathbf{d}_{j-1}, \mathbf{l}_{{att}_j}, \mathbf{r}_{{att}_j}\big] + \mathbf{b}_{dec})$$
\noindent where $\mathbf{W}_{dec}$ and $\mathbf{b}_{dec}$ are model parameters, $\mathbf{d}_{j-1}$ is the previous decoder hidden state, and $\mathbf{l}_{{att}_j}$ and $\mathbf{r}_{{att}_j}$ are the resulting attention vectors over the left and right segments, respectively, of encoder hidden states $\mathbf{h}_1 \dots \mathbf{h}_n$. These two segments of the input are defined by index $p$, which is initialized to the beginning of the sentence
and moves one position to the right each time a $\sh$ transition is applied. Therefore, $\mathbf{l}_{{att}_j}$ and $\mathbf{r}_{{att}_j}$ are computed at timestep $j$ as:
$$\mathbf{l}_{{att}_j} = \sum_{i=1}^{p}\alpha_{ij}\mathbf{h}_i,\ \ \  \mathbf{r}_{{att}_j} = \sum_{i=p+1}^{n}\alpha_{ij}\mathbf{h}_i,$$
where \ \ $\alpha_{ij} = \frac{exp(\beta_{ij})}{\sum_{k=1}^{n} exp(\beta_{kj})}$ \ \ and
$$\beta_{ij} = \mathbf{U}^T\mathbf{tanh}(\mathbf{W}_{att}\big[\mathbf{h}_i;\mathbf{d}_{j-1}\big]+\mathbf{b}_{att})$$ 
\noindent Then, the current token $\mathbf{y}_j$ is predicted from $\mathbf{d}_j$ as:
$$p(\mathbf{y}_j|\mathbf{d}_j) = \mathbf{softmax}(\mathbf{W}_{pred}*\mathbf{d}_j+\mathbf{b}_{pred}),$$
\noindent where $\mathbf{W}_{att}$, $\mathbf{b}_{att}$, $\mathbf{W}_{pred}$ and $\mathbf{b}_{pred}$ are parameters. In Figure~\ref{fig:network}, we graphically describe the neural architecture.

Note that current state-of-the-art transition-based parsers, which rely on stack-LSTMs to represent the stack structure, are also implemented under the framework by \citet{Dyer2015} and, therefore, our approach can be fairly compared to them in terms of accuracy and speed.

\label{sec:detatt}
\paragraph{Deterministic Attention} Previous work
\cite{Kamigaito2017,Ma2017,Liu2018} claims that using deterministic attention mechanisms instead of the standard probabilistic variant 
leads to accuracy and speed gains.
We propose a simple and effective procedure to implement deterministic attention in the architecture by \citet{LiuS2S17}, substantially reducing the time consumed by the decoder to predict the next token.

 Apart from dividing the sequence of encoder hidden states into segments, \citet{LiuS2S17} provide 
 explicit alignment between the input 
 word sequence
 and the output 
 transition sequence
 by keeping the index $p$ that indicates a correspondence between input words and $\sh$ transitions. 
 This information can be used to force the model to focus on those encoder hidden states that are more informative for 
 decoding
 at each timestep, avoiding going through the whole input to compute the attention vector,
 and thus considerably reducing decoding time.
 
 To 
 gain some insight on
 what 
 input words
 are most relevant, we study on the dev set the attention values assigned by the model to each encoder hidden state and the frequency with which each 
 of them achieves the highest value at each timestep. Surprisingly, we found out that, for the top-down parser, almost 90\% of the time the highest attention values were assigned to the words in positions $p$ and $p+1$
 by a wide margin.
 For the in-order parser,
 words in 
 those positions
 also received considerable attention values, but they were determinant only 75\% of the time.
 Following these results, we propose a computation of $\mathbf{l}_{{att}_j}$ and $\mathbf{r}_{{att}_j}$ where only the encoder hidden states in the rightmost position ($p$) of the left segment and in the leftmost position ($p+1$) of the right segment are considered:
$$\mathbf{l}_{{att}_j} = \beta_{pj}\mathbf{h}_p,\ \ \  \mathbf{r}_{{att}_j} = \beta_{p+1j}\mathbf{h}_{p+1}$$
\noindent This 
change
avoids calculating the weight $\alpha_{ij}$ for each encoder hidden state, 
as 
needed 
in probabilistic attention.
Attention vectors are computed in constant time, notably reducing running time while keeping 
the accuracy, as shown in our experiments.

\section{Experiments}
\label{sec:results}
We test the proposed approaches on 
the PTB treebank \cite{marcus93} with standard splits.\footnote{Settings are detailed in Appendix~\ref{sec:appendixDS}.}

\begin{table}[h]
\begin{small}
\centering
\begin{tabular}{@{\hskip 0.1pt}l@{\hskip 0.1pt}c@{\hskip 5.5pt}cc@{\hskip 0.5pt}}
Parser & Beam & Strat & F1 \\
\hline
\textbf{Transition-based} & & &  \\
\ \ \cite{Cross2016A} & n & bu & 90.0  \\
\ \ \cite{Cross2016B} & n & bu & 91.3  \\
\ \ \cite{Liu2017}  & n & bu & 91.3  \\
\ \ \defcitealias{nonbinary}{Fern\'andez-G and G\'omez-R, 2019}\citepalias{nonbinary}$^*$ & n &  bu & 91.7\\
\ \ \cite{Dyer2016}$^*$ & n & td & 91.2    \\
\ \ \defcitealias{FerGom2018}{Fern\'andez-G and G\'omez-R, 2018}\citepalias{FerGom2018}$^*$ & n & td & 91.7    \\
\ \ \cite{Liu2017}$^*$  & n & in & 91.8   \\
\ \ \defcitealias{FerGom2018}{Fern\'andez-G and G\'omez-R, 2018}\citepalias{FerGom2018}$^*$  & n & in & 92.0   \\
\ \ \cite{Zhu13} & y & bu & 90.4  \\
\ \ \cite{Watanabe15} & y & bu & 90.7  \\
\ \ \cite{Liu2017B} & y & bu & 91.7    \\
\ \ \cite{FriedK18} & y & in & \textbf{92.2}    \\
\textbf{Seq2seq} & & &  \\
\ \ \cite{Vinyals2015} & y & td & 88.3  \\
\ \ \cite{Ma2017} & y & bu & 88.6  \\
\ \ \cite{Kamigaito2017} & y & td & 89.5  \\
\ \ \cite{Liu2018} & y & td & 91.2  \\
\ \ \cite{Suzuki2018} & y & td & 91.2  \\
\ \ \cite{LiuS2S17}$^*$ \textit{(baseline)} & n & td & 90.5  \\
\ \ \textbf{Top-down SH-RE w/ det. attention}$^*$ & n & td & 90.7  \\
\ \ \textbf{Enriched top-down SH-RE}$^*$ & n & td & 90.7  \\
\ \ \textbf{In-order SH-RE}$^*$ & n & in & 90.9  \\
\ \ \textbf{Enriched in-order SH-RE}$^*$ & n & in & \textbf{91.3}  \\
\ \ \ \ \ \ \ \ \textbf{w/o lexicalized SH transition}$^*$ & n & in & 91.2  \\
\ \ \ \ \ \ \ \ \textbf{w/ det. attention}$^*$ & n & in & 91.2  \\
\ \ \ \ \ \ \ \ \textbf{w/ beam-search}$^*$ & y & in & \textbf{91.6}  \\

\hline
\hline
\textbf{Chart-based} & &   \\
\ \ \cite{SternAK17} & n & bu & 91.8 \\
\ \ \cite{GaddySK18} & n & bu & 92.1 \\
\ \ \cite{KleinK18} & n & bu & \textbf{93.6} \\
\hline
\end{tabular}
\caption{Accuracy comparison on PTB test set with greedy ($n$) or beam-search ($y$) decoding and with different strategies followed to parse or to linearize the input sentence ($bu$=bottom-up, $td$=top-down and $in$=in-order). Systems marked with $^*$ are implemented under the same framework.
}

\label{tab:results}
\end{small}
\end{table}
Table~\ref{tab:results} compares parsing accuracy of all linearizations proposed in Section~\ref{sec:lin} 
to  state-of-the-art fully-supervised transition-based constituent  parsing models. The results show that our
enriched in-order linearization is the most suitable option 
implemented 
so far for seq2seq constituent parsing, outperforming all existing seq2seq approaches (even without beam-search decoding) and matching some transition-based models. 
We also demonstrate that 
the enriched top-down variant (equivalent to the bracketed \cite{Vinyals2015}'s linearization) 
outperforms
the regular top-down approach 
of
\citet{LiuS2S17}. A trend that can  also be seen in the in-order linearization, where the addition of more tokens (parametrized $\re$ and lexicalized $\sh$ transitions) to the vocabulary benefits model performance (a gain of 0.4 
F-score points), meaning that seq2seq models make use of this additional information. In fact, we analysed the average length of output sequences and noticed that enriched variants with larger vocabulary tend to produce shorter sequences. We hypothesize that the extra information 
is helping the model to better contextualize tokens in the sequence during training, minimizing the prediction of wrong tokens at decoding time. Finally, we extend the implementation by \citet{LiuS2S17} with 
10-beam-search decoding
and increase F-score by 0.3 points.

We also evaluate parsing speeds under 
the exact same 
conditions among our approach and the top-down \cite{Dyer2016} and in-order \cite{Liu2017} transition-based 
constituent 
parsers, implemented in the framework by \citet{Dyer2015}.\footnote{Please note that the implementation by \citet{Dyer2015} is not optimized for speed, but it can be used as a common framework to compare different approaches.}
Table~\ref{tab:speed}
shows
how the proposed deterministic attention technique 
doubles the speed of the baseline model, putting it on par
with stack-LSTM-based shift-reduce systems, which are considered one of the most efficient approaches 
for
constituent parsing. 
We can also see from Table~\ref{tab:results} that the presented mechanism is more beneficial in terms of accuracy for the top-down algorithm (increasing 0.2 points in F-score) 
than the in-order variant
(suffering a drop of 0.1 points in F-score), 
as could be expected 
from
our previous analysis of 
attention vectors.

Finally, at the bottom of Table~\ref{tab:results}, we show current state-of-the-art chart-based parsers. These approaches, while more accurate, are significantly slower than seq2seq and transition-based parsers, being less appealing for downstream applications where the speed is crucial.

\begin{table}
\begin{small}
\centering
\begin{tabular}{@{\hskip 0.1pt}lc@{\hskip 0.1pt}}
Parser & sent./s.  \\
\hline
\textbf{Transition-based} &  \\
\ \ \cite{Dyer2016} \textit{(top-down)} & 38.78    \\
\ \ \cite{Liu2017} \textit{(in-order)} & 33.34   \\
\textbf{Seq2seq} &  \\
\ \ \cite{LiuS2S17} \textit{(top-down SH-RE)} & 16.65  \\
\ \ \textbf{Top-down SH-RE w/ det. attention} & 37.93  \\
\ \ \textbf{Enriched in-order SH-RE} & 16.54  \\
\ \ \textbf{Enriched in-order SH-RE w/ det. attention} & 35.12 \\
\hline
\end{tabular}
\caption{Speed comparison on PTB test set. 
}

\label{tab:speed}
\end{small}
\end{table}

\section{Conclusion}
We present significant
accuracy and speed improvements
in seq2seq constituent parsing. The proposed linearization techniques can be used by any off-the-self seq2seq model without building a specific algorithm or structure. In addition, any advances in seq2seq neural architectures or pre-trained transformer-based language models \cite{BERT} can be directly used to enhance our approach.

\section*{Acknowledgments}
This work has received funding from the European
Research Council (ERC), under the European
Union's Horizon 2020 research and innovation
programme (FASTPARSE, grant agreement No
714150), from the ANSWER-ASAP project (TIN2017-85160-C2-1-R) from MINECO, and from Xunta de Galicia (ED431B 2017/01, ED431G 2019/01).

\bibliography{main,twoplanaracl}
\bibliographystyle{acl_natbib}

\appendix

\section{Appendices}
\label{sec:appendix}
\subsection{Top-down Transition System}
\label{sec:appendixTD}
 In the top-down transition system defined by \citet{Dyer2016}, parser configurations have the form {$c=\langle {\Sigma} , {B} \rangle$}, where $\Sigma$ is a \textit{stack} of constituents and $B$ is the \textit{buffer} that contains words from the input sentence. The top-down algorithm also provides three transitions (described in Figure \ref{fig:transitions}) that can be used on the stack and the buffer (that initially contains the whole unparsed sentence) to build the final constituent tree. Concretely: 
\begin{itemize}
    \item a \textit{Shift} transition is used to push words from the buffer into the stack,
    \item a \textit{Non-Terminal-X} transition to push a non-terminal node X into the stack,
    \item and a \textit{Reduce} transition to pop elements from the stack until a non-terminal node is found and create a new subtree with all these elements as its children, pushing this new constituent into the stack.
\end{itemize}

\subsection{In-order Transition System}
\label{sec:appendixIO}
\citet{Liu2017} define a transition system 
that builds a phrase structure tree in an in-order traversal order: 
the non-terminal node on top of the 
tree being built
is only considered after the first child 
node 
is completed in the stack, building each subtree in a bottom-up manner, but choosing the non-terminal node on top before the new constituent is reduced. 
This transition system has parser configurations with the stack-buffer form {$c=\langle {\Sigma} , {B} \rangle$} and uses the following actions (described in Figure \ref{fig:transitions2}):
\begin{itemize}
    \item a \textit{Shift} transition to move words from the buffer to the stack,
    \item a \textit{Non-Terminal-X} transition to push a non-terminal node X into the stack as long as the first child of the future constituent is on top of the stack,
    \item a \textit{Reduce} transition to pop all elements from the stack until the first non-terminal node is found, which is also popped together with the preceding element in the stack to build a new constituent with all of them as children of the non-terminal node,
    \item and, finally, a \textit{Finish} transition to terminate the parsing process.
\end{itemize}

The in-order transition system is a combination of the classic bottom-up and the new top-down algorithms, providing advantages of both of them: the access to information from partial parses from the bottom-up approach, and the non-local outlook of the top-down approach. 

\begin{figure}
\begin{tabbing}
\hspace{0.1cm}\=\hspace{0.7cm}\= \kill
\> Shift: 
\> \ \ \ \ \ \ \ $\langle {\Sigma}, {w_i | B}  \rangle
\Rightarrow \langle {\Sigma | w_i} , {B}  \rangle$\\[2mm]
\> NT-X:
\>  \ \ \ \ \ \ \ $\langle {\Sigma}, B \rangle
\Rightarrow \langle {\Sigma | X }, B  \rangle$ { }\\[2mm]
\> Reduce:
\>   \ \ \ \ \ \ \   $\langle {\Sigma} | X| s_k| \dots | s_0 , B, \rangle \Rightarrow \langle {\Sigma} | X_{s_k \dots  s_0} , B \rangle$

\end{tabbing}
\caption{Transitions available in a top-down transition system (NT-X = Non-Terminal-X).}
\label{fig:transitions}
\end{figure}

\begin{figure}
\begin{tabbing}
\hspace{0.1cm}\=\hspace{0.7cm}\= \kill
\> Shift: 
\> \ \ \ \ \ \ \ $\langle {\Sigma}, {w_i | B} , false \rangle
\Rightarrow \langle {\Sigma | w_i} , {B}, false  \rangle$\\[2mm]
\> NT-X:
\>  \ \ \ \ \ \ \ $\langle {\Sigma}, B, false \rangle
\Rightarrow \langle {\Sigma | s_0 | X }, B, false  \rangle$ { }\\[2mm]
\> Reduce:
\>   \ \ \ \ \ \ \   $\langle {\Sigma} | s_k| X| s_{k-1}| \dots | s_0 , B, false \rangle$\\
\>  \ \ \ \ \ \ \  \hspace{2.5cm}  $\Rightarrow \langle {\Sigma} | X_{s_k \dots  s_0} , B, false \rangle$ { } \\[2mm]
\> Finish:
\>  \ \ \ \ \ \ \ $\langle {\Sigma}, B, false \rangle
\Rightarrow \langle {\Sigma}, B , true \rangle$ { }
\end{tabbing}
\caption{Transitions available in an in-order transition system (NT-X = Non-Terminal-X).}
\label{fig:transitions2}
\end{figure}

\begin{table}
\begin{small}
\centering
\begin{tabular}{@{\hskip 0pt}lc@{\hskip 0pt}}
\hline
\textbf{Hyper-parameters} & \\
\hline
BiLSTM encoder layers & 2 \\
BiLSTM encoder input size & 100 \\
BiLSTM encoder hidden size & 200 \\
LSTM decoder layers & 1 \\ 
LSTM decoder hidden size & 400 \\
POS tag embedding dimension & 6 \\
Pretrained word embedding dimension & 100\\ 
Word embedding dimension & 64\\
Label embedding dimension & 20\\
Action embedding dimension & 40\\
Attention hidden size & 50 \\
Initial learning rate & 0.001 \\
$\beta_1$, $\beta_2$ & 0.9 \\
 $\lambda$ & 10$^{-6}$\\
 \hline
\multicolumn{1}{c}{}\\
\end{tabular}
\centering
\setlength{\abovecaptionskip}{4pt}
\caption{Model hyper-parameters.}
\label{tab:hyper}
\end{small}
\end{table}

\subsection{Data and Settings}
\label{sec:appendixDS}
Following common practice, we test the proposed approaches on 
the Wall Street Journal sections of the English Penn Treebank \cite{marcus93} with standard splits: sections  2-21 are used as training data, section 22 for development and section 23 for testing.

We adopt stochastic gradient descent with Adam \cite{Adam} and hyper-parameter selection as \cite{LiuS2S17}, detailed in Table~\ref{tab:hyper}. In addition, we use  predicted POS tags and pre-trained word embeddings (generated on the AFP portion of English Gigaword) as \cite{Dyer2016,LiuS2S17,Liu2017}. 

All neural models are trained by minimizing the following cross-entropy loss objective with an $l_2$ regularization term:
$$\mathcal{L}(\theta)=-\sum_{i} \sum_{j}log\ p_{y_{ij}} + \frac{\lambda}{2}||\theta||^2$$
\noindent where $\theta$ is the set of parameters, $p_{y_{ij}}$ is the probability of the $j$th token in the $i$th training example given by the model and $\lambda$ is a regularization hyper-parameter. For further details about the neural architecture, the reader can refer to \cite{LiuS2S17}. 

For our executions, 
we report the average accuracy and speed over 3 runs with random initialization and on a single CPU core.

\end{document}